\documentclass{article}

\PassOptionsToPackage{numbers, compress}{natbib}

\usepackage[preprint]{neurips_2025}
\usepackage[utf8]{inputenc} 
\usepackage[T1]{fontenc}    
\usepackage{fix-cm}         
\usepackage{url}            
\usepackage{booktabs}       
\usepackage{amsfonts}       
\usepackage{nicefrac}       
\usepackage{microtype}      
\usepackage{xcolor}         
\usepackage{hyperref}
\usepackage{bm}
\usepackage{ulem}
\usepackage{multirow}
\usepackage{fontsize}
\usepackage{graphicx}
\usepackage{wrapfig}
\usepackage{subfigure}
\usepackage{amsmath}
\usepackage{algorithm2e}
\def\model{REEL}

\title{Bridging the Domain Gap in Equation Distillation with Reinforcement Feedback}

\author{%
\textbf{Wangyang Ying}\textsuperscript{1}, 
\textbf{Haoyue Bai}\textsuperscript{1}, 
\textbf{Nanxu Gong}\textsuperscript{1}, 
\textbf{Xinyuan Wang}\textsuperscript{1}, 
\textbf{Sixun Dong}\textsuperscript{1}, \\
\textbf{Haifeng Chen}\textsuperscript{2}, 
\textbf{Yanjie Fu}\textsuperscript{1} \\
\textsuperscript{1}Arizona State University, Tempe, USA \\
\textsuperscript{2}NEC Laboratories America, Inc., Princeton, USA \\
\texttt{\{wying4, haoyuebai, xwang735, sixundong, yanjie.fu\}@asu.edu}, \\ \texttt{haifeng@nec-labs.com}
}

\begin{document}

\maketitle

\begin{abstract}
    The data-to-equation (Data2Eqn) task aims to discover interpretable mathematical equations that map observed values ($\mathbf{X}$) to labels ($y$), offering physical insights and broad applicability across academic and industrial domains. 
    Genetic programming and traditional deep learning-based approaches suffer from search inefficiency and poor generalization on small task-specific datasets.
    Foundation models showed promise in this area, but existing approaches suffer from: 
    1) They are pretrained on general-purpose data distributions, making them less effective for domain-specific tasks; 
    and 2) their training objectives focus on token-level alignment, overlooking mathematical semantics, which can lead to inaccurate equations. 
    To address these issues, we aim to enhance the domain adaptability of foundation models for Data2Eqn tasks.
    In this work, we propose a reinforcement learning-based finetuning framework that directly optimizes the generation policy of a pretrained model through reward signals derived from downstream numerical fitness. Our method allows the model to adapt to specific and complex data distributions and generate mathematically meaningful equations. Extensive experiments demonstrate that our approach improves both the accuracy and robustness of equation generation under complex distributions. 
\end{abstract}
\vspace{-0.4cm}
\section{Introduction}
\vspace{-0.2cm}
The data-to-equation (Data2Eqn) task, also known as symbolic regression, identifies the optimal equation that maps observed values ($\mathbf{X}$) to labels ($y$) by exploring the mathematical expression space. Unlike other modeling approaches that rely on predefined functional forms, this method learns interpretable equations directly from data and offers clear physical insights.
Data2Eqn tasks have significant applications across various domains; for instance, they are used to discover physical laws, optimize industrial designs, analyze economic causal relationships, extract biological mechanisms from genomic data, etc~\citep{wang2019symbolic,angelis2023artificial,cranmer2020discovering,can2011comparison,claveria2017assessment,la2023flexible}. These tasks can model complex systems, support automated scientific discovery, and ensure strong model interpretability.

\noindent Previous works are divided into three classes:
1) Genetic Programming (GP) methods~\citep{GP,GP2,GP3};
2) Traditional deep learning approaches~\citep{petersen2019deep,kim2020integration,zhang2023deep,d2022deep};
3) Foundation models~\citep{biggio2021neural,kamienny2022end,shojaee2023transformer,holt2023deep,valipour2021symbolicgpt}.
GP methods optimize equations iteratively through genetic operations such as crossover and mutation. Although it can achieve strong performance, it suffers from high search complexity, sensitivity to hyperparameters, low computational efficiency, and a tendency to get stuck in local optima.
Traditional deep learning methods rely on training with task-specific data and lack generalization capability. They often require retraining when the data distribution shifts, which results in high adaptation costs and limited transferability.
Foundation models are pretrained on large-scale datasets to learn transferable knowledge and robust representations. They have been widely adopted for Data2Eqn tasks. The core idea is to encode observed values and labels(e.g., $\mathbf{X}$ and $y$) into latent embeddings and decode them into equations represented as token sequences. Although these models capture general equation patterns (e.g., variable compositions and numerical patterns), they don't capture the data distribution of specific tasks and may introduce domain-inconsistent biases (a.k.a., negative transfer).
Therefore, this research gap motivates our study, which focuses on effectively adapting foundation models to specific domains (e.g., physics) to extract accurate equations.

\noindent The Data2Eqn task faces two main challenges in adapting foundation models to specific datasets:
1) Data distribution mismatch: The foundation model is pretrained on large-scale general-purpose data randomly sampled from a normal distribution, which does not reflect the data characteristics commonly found in scientific domains such as physics, biology, and materials science. This mismatch can lead to negative transfer.
2) Training objective mismatch: The pretraining objective adopts a token-level cross-entropy loss inspired by next-token prediction in text generation. This loss focuses on token-level alignment rather than mathematical semantics, which may result in equations that are structurally correct but numerically inaccurate, resulting in negative transfer.

\noindent\textbf{Our Insight: Through iterative reward-driven adaptation, reinforcement learning (RL) transfers foundation models' knowledge to domain-specific patterns beyond their pretraining scope.} 
We propose an RL–based finetuning framework to enhance the equation generation ability of foundation models in specific and complex domains.
To address challenge 1, we randomly sample multiple subsets from the domain-specific dataset, treating each as an independent equation generation task. Our framework directly optimizes the model’s generation policy by interacting with the task environment, enabling alignment with both the mathematical meaning and the data distribution of the target domain.
To address challenge 2, the model generates candidate equations under its current policy and receives reward signals. The reward measures how well each equation fits the target data in terms of mathematical semantics, rather than token-level similarity.
To further stabilize training and retain valuable prior knowledge, we incorporate a KL divergence regularization term between the finetuned and pretrained models. This constrains policy updates to stay close to the pretrained distribution while improving performance through environment-specific feedback. We employ a policy gradient approach to iteratively update the generation strategy, allowing the model to progressively learn the structural and distributional patterns of the target domain.

\noindent\textbf{Summary of our contributions:}
1) \underline{\textit{RL-based Policy Adaptation:}} We introduce an RL–based finetuning framework into pretrained Data2Eqn models, where reward signals guide the generation policy to adapt to complex domain-specific data distributions, thereby improving equation generation.
2) \underline{\textit{Numerical-Semantic Reward Design:}} We adopt a straightforward RL objective by using numerical fitness as reward signals. This allows the generation policy to receive feedback on the mathematical quality of equations.
3) \underline{\textit{Empirical Validation on Data2Eqn:}} Extensive experiments demonstrate that our method significantly enhances the equation generation ability and robustness of foundation models under complex data distributions, validating the effectiveness of RL in the Data2Eqn task.
\vspace{-0.4cm}
\section{Preliminaries and Problem Statement}
\vspace{-0.3cm}

\noindent\textbf{Data to Equation.}
The data-to-equation task, also known as symbolic regression, aims to uncover an explicit mathematical expression that explains the relationship between inputs and outputs.  Formally, given a dataset $\mathcal{D} = (\mathbf{X}, y)$, where $\mathbf{X}$ is an input predictor matrix, and $y$ is a target variable (labels), the objective is to find an interpretable equation $f$ such that $y = f(\mathbf{X})$.

\noindent\textbf{Foundation Data to Equation Models with Transformer Backbones.} 
The study in~\cite{biggio2021neural} introduced a Transformer-based pretraining framework to tackle the data-to-equation problem. 
\emph{Training Data: }
The foundation model is trained on large-scale synthetic data to learn how to map predictor-target pairs to an equation.
To prepare the training data, many input predictor matrices  $\{\mathbf{X}_i\}_{i=1}^N$ are randomly sampled from a Gaussian distribution, where $N$ is the number of matrices. 
For each matrix $\mathbf{X}_i$, a corresponding equation $f_i$ is generated by combining predictors and operators selected from a predefined set of mathematical operations.
The target variable $y_i$ is computed by evaluating the equation $f_i(\mathbf{X}_i)$. 
In this way, a training dataset of $N$ triplets $\{(\mathbf{X}i, y_i, f_i)\}_{i=1}^N$ is prepared for pretraining.
\noindent\emph{Model Architecture:} The foundation model is based on a Transformer encoder-decoder design: the encoder processes $(\mathbf{X}_i, y_i)$ pairs and converts them into latent embeddings; the decoder decodes the embeddings to generate the corresponding symbolic equation $f_i$.

\noindent However, synthetic pretraining data doesn't necessarily match the distributions of real-world domains (e.g., healthcare,  materials science), and the foundation model suffers from negative transfer. The distributional mismatch leads to poor performance when the foundation model is applied to domain-specific datasets, limiting its practical deployment.
Therefore, it is needed to develop a tool to finetune and adapt a foundation equation distillation model to domain-specific datasets.

\noindent\textbf{Problem Statement: Reinforcement Enhanced Equation Learning for Foundation Data2Eqn Models (\model)}. 
Formally, given a pretrained Data2Eqn foundation model and a target dataset, the goal is to develop a fine-tuning framework that aligns the equation generation ability of the foundation model with the domain knowledge of the target dataset.

\vspace{-0.2cm}
\section{Reinforcement Task-optimal Adaptation of Foundation Data2Eqn Models}
\vspace{-0.2cm}
\begin{figure*}[th]
    \centering
    \includegraphics[width=1\linewidth]{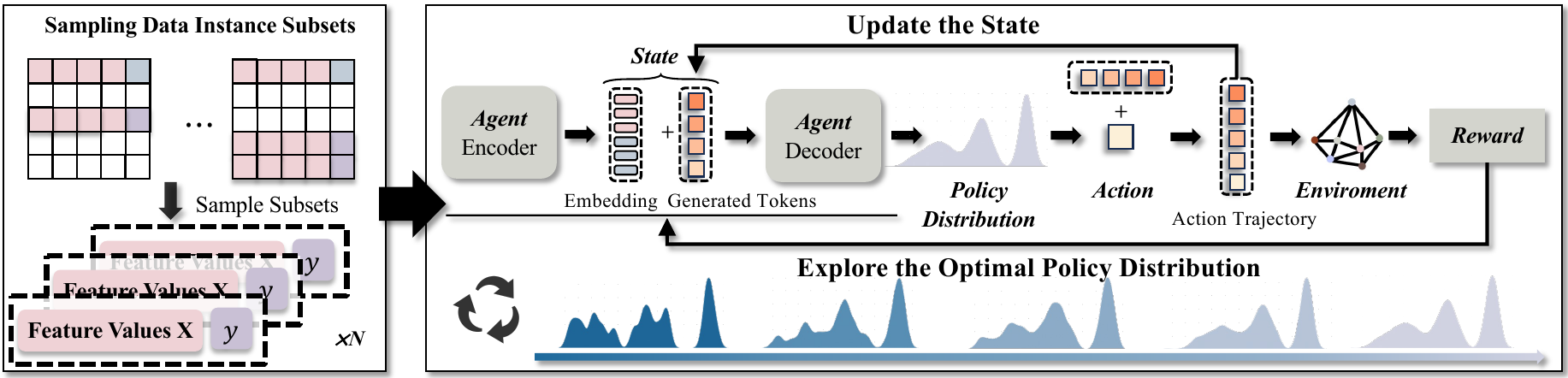}
    \caption{Framework Overview. Given a domain-specific dataset, we first sample diverse instance subsets to form multiple trajectory training sets. For each subset, an agent (a pretrained Data2Eqn model) iteratively generates equation tokens, receives numerical rewards based on equation-level fitness, and updates its policy via reinforcement learning. This process explores the optimal policy distribution and aligns generation with domain-specific mathematical semantics.}
    \label{fig:framework}
    \vspace{-0.2cm}
\end{figure*}
\subsection{Overview of the Proposed Solution}
\vspace{-0.2cm}
Given a domain-specific tasking dataset with features and labels, and a pretrained Data2Eqn foundation model, 
the objective is to finetune the pretrained model on the tasking dataset to generate the best-fitting equation that describes the mapping from features to labels. The underlying AI task is task-optimal adaptation from a foundation model to a domain dataset under the context of equation distillation.
\textbf{Figure \ref{fig:framework}} shows the two steps of our framework. \textbf{Step 1: sampling data instance subsets} is to sample a large number of data subsets from the instances (rows) of the tasking dataset. 
Sampling instance subspaces from tasking data can allow us to finetune the foundation model to learn a robust and deserving subspace-driven feature-label relationship within the same tasking data. 
\textbf{Step 2: reinforcement task-optimal adaptation} is to leverage reinforcement learning from feedback in the loop to finetune the pretrained Data2Eqn model over diverse data instance subsets of the tasking data, so that the finetuned model can generate the best-fitting equation of the tasking data. 

\subsection{Tasking Data Instance Subset Sampling for Data-centric Foundation Model Adaptation}
\vspace{-0.2cm}
Foundation Data2Eqn models are widely trained on generic tabular data. 
Our goal is to adapt foundation Data2Eqn models to a specific tasking dataset via merging reinforcement finetuning techniques. 
In reinforcement learning, we need diverse different trajectories during training to effectively explore different states in the environment, avoid getting stuck in suboptimal behaviors, learn diverse strategies that generalize well, and get accurate estimates of expected rewards for reliable policy improvements.
Under the context of Data2Eqn, one reinforcement exploration trajectory can be seen as the exploration process that a reinforcement agent learns to derive the best-fitting equation from a data sample (e.g., subspace, subset, copy) of the tasking data. 
From a data-centric perspective, a reinforcement trajectory corresponds to a sample of the tasking data. 
Diversifying trajectories in reinforcement learning plays a similar role to bootstrap sampling in Random Forests: both introduce variability to reduce overfitting and improve robustness. 

\noindent Inspired by this computing thinking, in Step 1, we propose a data-centric perspective: sampling diverse tasking data subspaces for diversifying trajectories, to better perceive the relational structures of tabular data and achieve more generic and reliable finetuning of the foundation Data2Eqn model. 
In particular, given the tabular tasking data $\mathcal{D} = (\mathbf{X}, y)$ consisting of multiple feature columns and one label column, where each row represents an instance (data point). 
We randomly sample a subset of instances with replacement, denoted by $(\mathbf{X}_i, y_i)$,  as the Data2Eqn training subset data of one reinforcement trajectory. 
We sample multiple times to generate a collection of finetuning trajectory training subsets $\mathcal{T} = (\mathbf{X}_i, y_i)_{i=1}^N$, where $N$ is the number of trajectory training subsets. 

\noindent These finetuning trajectory training subsets cover diverse subspaces and corresponding X-y relationship structures in corresponding subspaces of the tasking data, thus allowing reinforcement agents to explore different states of the environment. Moreover,  transformer-based Data2Eqn foundation models have input length limits due to the exponential computational and memory demands of the self-attention mechanism as tabular sequence length increases, such instance subspaces as finetuning trajectory training data can ensure the compliance of input limits of Transformer based foundation model to advance the availability and practical generalization of our method for all kinds Transformer-based foundation models. 

\vspace{-0.3cm}
\subsection{Reinforcement Finetuning for Learning-centric Foundation Model Adaptation}
\vspace{-0.2cm}
We first design reinforcement Data2Eqn finetuning with a single trajectory training subset, then extend reinforcement Data2Eqn finetuning to multiple trajectory training subsets. 

\vspace{-0.3cm}
\subsubsection{Reinforcement Finetuning with Single Trajectory Training Subset}
\vspace{-0.2cm}
The emergence of DeepSeek demonstrates the success of reinforcement finetuning in finetuning the reasoning capabilities of LLMs. 
Similarly, we find that reinforcement learning shows significant potential for adapting foundation Data2Eqn models to specific tasking data, as reinforcement intelligence can achieve foundation model finetuning through task-specific feedback, explore beyond limited training distributions, and make foundation models more steerable. 
We thus propose to analogize and formulate the foundation Data2Eqn model finetuning on a tasking dataset as a reinforcement learning process. Reinforcement learning has six key elements: agents, states, actions, environment, rewards, and policies. We design a new analogical reinforcement learning based reformulation for Data2Eqn foundation model finetuning: 

\noindent\textbf{The Agent.} We treat the transformer-based encoder-decoder model as an agent that learns to refine the equation distillation through reinforcement learning. The agent starts with the foundation model~\cite{biggio2021neural} pretrained on massive tabular data and corresponding equations to acquire general data relationship understanding and equation distillation abilities. The agent isn't a static model but rather the foundation model undergoing finetuning through reinforcement to update its parameters to improve its equation distillation capability.

\noindent\textbf{The Environment.} 
The environment corresponds to the Data2Eqn task. The agent interacts with the task by sequentially generating equation tokens and receives equation-level feedback based on the evaluation of the complete equation on the training data. At each decoding step, the environment updates the state by appending the newly generated token to the existing sequence, forming a new prefix that progresses the next decoding step. This evolving sequence, together with the input data, defines the state for subsequent token prediction. 
When all the tokens of an equation are generated, the environment parses the tokens into a symbolic equation and verifies equation validity. 

\noindent\textbf{The State of Environment.}  
The state of the environment represents the current context in which the equation distillation model is generating equation tokens. The state representation includes: i) the input data, which is the inputs of the encoder and will be decoded into an equation; ii) the sequence of equation tokens generated so far in response to the input.
Formally, let $t$ be current decoding step, the state representation vector $\mathbf{s}_t$ is given by $\mathbf{s}_t = [\mathbf{e}, a_{<t}]$, where  $\mathbf{e}$ is the encoder’s output embedding of input data, and  $a_{<t} = (a_1, ..., a_{t-1})$ is the equation token sequence generated so far.

\noindent\textbf{The Policy Function: A Mapping from State to Token.}  
The policy function corresponds to the decoder’s token-level probability distribution over the equation vocabulary, conditioned on the current state at each decoding step.
To this end, the input data and previously generated tokens are processed as the current state to compute logits over the vocabulary, which are subsequently normalized into a probability distribution using softmax.
Formally, let $\mathbf{s}_t$ be the current state at $t$, $\psi$ is the decoder, $V$ is the token vocabulary, the probability distribution over the equation token vocabulary when generating the current token is given by: $ \pi_\theta(a_t|\mathbf{s}_t) = softmax(\mathbf{s}_t),$
where $\theta$ is the parameter of the model.

\noindent\textbf{The Actions.} 
An action is to generate the next equation token given the state, which is guided by the learned policy and the goal of maximizing rewards associated with the generated equation tokens. 
In particular, in an action, the token with the highest predicted probability is selected:
\begin{equation}
    a_t = \arg\max_{a \in V} \pi_\theta(a|\mathbf{s}_t),
\end{equation}
where $V$ is the equation token vocabulary, $a\in V$ is an equation token, $t$ is current timestep, $\mathbf{s}_t$ is current state. 
The decoding process begins with a special start-of-sequence token <sos> and terminates either when the end-of-sequence token <eos> is generated or when the sequence reaches a predefined maximum length, along with a generated equation token sequence: $a_{1:T} = (a_1, ..., a_T)$, where $T$ is the max length of an equation. 

\noindent\textbf{Smoothed Reward Function:} 
The reward is a numerical signal that indicates the numerical fitness of the generated equation on training data with respect to the ground-truth labels. 
After the Data2Eqn model generates an equation token sequence, we propose a two-step method: 1) We compute  the coefficient of determination between the predicted labels by the generated equation and the benchmark labels of the training data as a reward score: 
\begin{equation}
    R^2(y, \tilde{y}) = 
    \begin{cases}
    1 - \frac{\Sigma_{i-1}^n(y_i - \tilde{y}_i)^2}{\Sigma_{i=1}^n(y_i - \bar{y}_i)^2},, \text{if the equation is valid} \\
    -1, \text{otherwise} \\
    \end{cases},
\end{equation}
where $\tilde{y}$ is equation-predicted label, and $y$ is the groud-truth.  
2) We transform the raw $R^2$ score using a smoothed sigmoid-based function:
$
    r = \frac{2}{1 + \exp(-R^2)} -1
$
to robustify reward quantification and protect against outliers. 

\begin{wrapfigure}{r}{0.25\textwidth}
    \centering
    \vspace{-0.5cm}
    \includegraphics[width=1\linewidth]{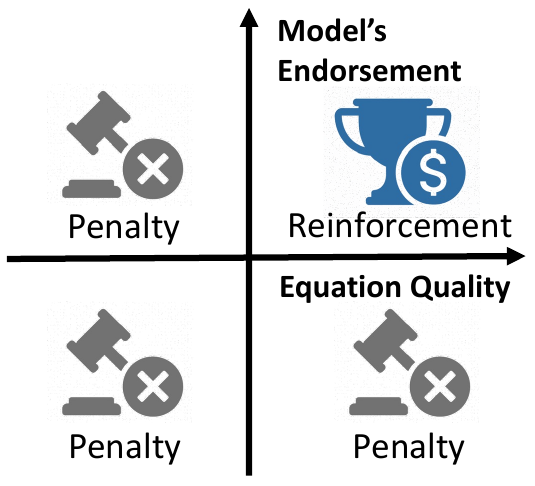}
    \caption{Reward assignment: we reinforce it when the finetuned model generates high-quality equations.}
    \vspace{-0.3cm}
    \label{fig:policy}
\end{wrapfigure}
\noindent\textbf{Policy Learning.} 
\textbf{Figure~\ref{fig:policy}} shows four modeling intuitions to guide reward assignment:
1) High-quality equations endorsed by the finetuned model should be reinforced;
2) High-quality equations endorsed by the foundation model suggest that the finetuned model fails to capture valuable patterns, and thus should be penalized; 
3) Low-quality equations endorsed by the finetuned model indicate poor finetuning and should be penalized to discourage undesirable generation;
4) Low-quality equations endorsed by the foundation model indicate limited capability in the foundation model, and no reward should be assigned in such cases.

\noindent\underline{\textit{Leveraging Endorsement Confidence for Clipped Reward Reweighing.}}
To model the intuitions, we compute the ratio between the finetuned model's and the foundation model's generative likelihoods, which serves as a measure of how much more confidently the finetuned model endorses a generated equation compared to the foundation model. 
Formally, let $a_{1:T}$ be a generated equation with $T$ tokens, $t$ indexes the tokens of the equation, $a_t$ is the t-th token, $\mathbf{s}_t$ is the t-th state representation,  $\pi_\theta(a_t|\mathbf{s}_t)$ is the probability of the t-th token under the policy of the finetune model, $\pi_\theta^{old}(a_t|\mathbf{s}_t)$ is the probability of the t-th token under the policy  of the foundation model, the relative endorsement confidence of the finetune model compared with the foundation model is given by:
\begin{equation}
    \rho = \exp (\frac{1}{T} \Sigma_{t=1}^T \log \frac{\pi_\theta(a_t|\mathbf{s}_t)}{\pi_{\theta}^{old}(a_t|\mathbf{s}_t)}), 
\end{equation}
where the logarithm $log$  transforms multiplications into summations to simplify calculations and improve numerical stability.
We thereafter employ a reward reweighing technique: multiplication of reward and relative endorsement confidence to unify and quantify the modeling intuitions as one score, denoted by $\rho\cdot r$, where $r$ is the reward of an equation.
Besides, we exploit a clipped surrogate function to restrict reward reweighing and prevent excessive incentivization or penalization, given by:
\begin{equation}
    \mathcal{L}_{clip} = -\min (\rho \cdot r, \quad \text{clip} (\rho, 1-\epsilon, 1+\epsilon) \cdot r).
\end{equation}
where the clip function limits a value to a specified range, and $\epsilon$ is a constant hyperparameter. 

\noindent\underline{\textit{Incorporating Policy Stability Regularization.}} 
Finetuning means adapting the foundation Data2Eqn model to a new tasking dataset. However, it is essential to constrain the finetuned policy from deviating too far from the foundation policy, avoid catastrophic forgetting, and ensure generalizability. 
We introduce a Kullback-Leibler divergence regularization term that is computed by comparing the foundation decoder distribution and the finetuned decoder distribution at each token decoding step:
\begin{equation}
    \mathcal{L}_{kl} = \frac{1}{T} \Sigma_{t=1}^T \Sigma_{a \in V} \pi_{\theta_{old}}(a|\mathbf{s}_t)[\log \pi_{\theta_{old}}(a|\mathbf{s}_t) - \log \pi_\theta(a|\mathbf{s}_t)].
\end{equation}

\noindent\underline{\textit{Finally}}, we integrate both clipped reward reweighing and policy stability regularization as reinforcement finetuning optimization objective: $\mathcal{L} = \mathcal{L}_{clip} + \beta \cdot \mathcal{L}_{kl}$, where $\beta$ is a balancing coefficient.  We optimize the policy $\pi_\theta$ via gradient descent using $\nabla_\theta \mathcal{L}$.

\subsubsection{Reinforcement Finetuning in Multiple Trajectory Training Subsets}
We have designed the reinforcement finetuning under one exploration trajectory with one trajectory training subset. 
Robust reinforcement finetuning needs diverse exploration trajectories and various training subsets to cover different data subspaces and relationship structures, and explore different states of the environment. 
Our idea is to iterate the reinforcement finetuning with a single trajectory training subset (Section 3.3.1) on multiple trajectory training subsets prepared in Section 3.2. The \textbf{Algorithm~\ref{alg:twoloop}} in \textbf{Appendix~\ref{appendix:algo}} provides an example to illustrate the two-loop finetuning process.

\vspace{-0.2cm}
\section{Experimental Results}
\vspace{-0.2cm}

We present intensive experimental results with public benchmarks on data-to-equation to evaluate the effectiveness and robustness of our method.
\vspace{-0.2cm}
\subsection{Experimental Settings}
\vspace{-0.2cm}
\noindent\textbf{Evaluation Metrics.} 
1) \textbf{$\bm{R^2}$ score:} $R^2 = 1-\frac{\Sigma_{i=1}^n(\Tilde{y}_i-y_i)^2}{\Sigma_{i=1}^n(\Tilde{y}_i-\bar{y})^2}$, where $R^2$ measure the fitness accuracy of the generated equations. 
2) \textbf{$\bm{R^2 > 0.99}$:} For a given domain-specific dataset (e.g., Feynman), $R^2 > 0.99$ denotes the proportion of equations achieving an $R^2$ score larger than 0.99. For example, if 80 out of 119 equations in the Feynman datasets reach this threshold, then $R^2 > 0.99 = \frac{80}{119} = 0.672$. 
3) \textbf{Data splitting:} We split the datapoints of each equation into training and testing datapoints at a ratio of 75\% and 25\%, respectively. 
The finetuning datasets are sampled from the training datapoints, while the test datapoints are used to evaluate model performance.

\noindent\textbf{Datasets.} We used three widely used standard benchmark datasets of the real world: 1) Feynman~\cite{udrescu2020aifeynman20paretooptimal}, 2) Strogatz~\cite{la2016inference}, and 3) Black-box~\cite{lacava2021contemporarysymbolicregressionmethods}. More details are provided in \textbf{Appendix~\ref{appendix:exp_set}}.

\noindent\textbf{Implementation Details.}
We leverage the state-of-the-art open-source End-to-End (E2E) model~\cite{kamienny2022end} as the pretrained transformer backbone. This selection is due to the public availability of E2E's model architecture, pretrained weights, and output logits through the Facebook symbolic regression library and its associated repository\footnote{\url{https://github.com/facebookresearch/symbolicregression}}. We describe the details of implementation in \textbf{Appendix~\ref{appendix:exp_set}}. 

\noindent\textbf{Baselines.}
We compare {\model} with the backbone E2E and 15 various SRBench algorithms\footnote{\url{https://github.com/cavalab/srbench}} to evaluate our proposed method. We describe the details of baselines in \textbf{Appendix~\ref{appendix:exp_set}}.

\setlength{\tabcolsep}{1.8mm}{
\begin{table*}[t]
\centering
\caption{Performance compared with the foundation model E2E~\cite{kamienny2022end}. $R^2 > 0.99$ indicates the proportion of discovered equations on the dataset where the test set achieves $R^2 > 0.99$. $\bar{R^2}$ represents the average test set performance across all discovered equations.}
\vspace{0.2cm}
\begin{tabular}{@{}clcccccccc@{}}
\toprule\toprule
\multirow{2}{*}{Model} &  & \multicolumn{2}{c}{Feynman}   &  & \multicolumn{2}{c}{Strogatz}  &  & \multicolumn{2}{c}{Black-box} \\ \cmidrule(lr){3-4} \cmidrule(lr){6-7} \cmidrule(l){9-10} 
                       &  & $\uparrow R^2 > 0.99$  & $\uparrow \bar{R^2}$     &  & $\uparrow R^2 > 0.99$ & $\uparrow \bar{R^2}$     &  & $\uparrow R^2 > 0.99$ & $\uparrow \bar{R^2}$     \\ \midrule
E2E-Beam               &  & 0.798                & 0.9727 &  & 0.286                & 0.7591 &  & 0.000                & 0.6968 \\
E2E-Sampling           &  & 0.815                & 0.9730 &  & 0.357                & 0.8156 &  & 0.000                & 0.7225 \\ \midrule
{\model}-Beam          &  & 0.891                & 0.9833 &  & 0.785                & 0.9819 &  & 0.053                & 0.7985 \\ 
{\model}-Sampling                   &  & \textbf{0.908}                & \textbf{0.9890} &  & \textbf{0.857}                & \textbf{0.9932} &  & \textbf{0.088}                & \textbf{0.8692} \\ \bottomrule \bottomrule
\end{tabular}
\label{exp:rl_comparison}
\end{table*}}

\begin{figure}[t]
  \centering
  \subfigure[Feynman]{
    \includegraphics[width=0.9\textwidth]{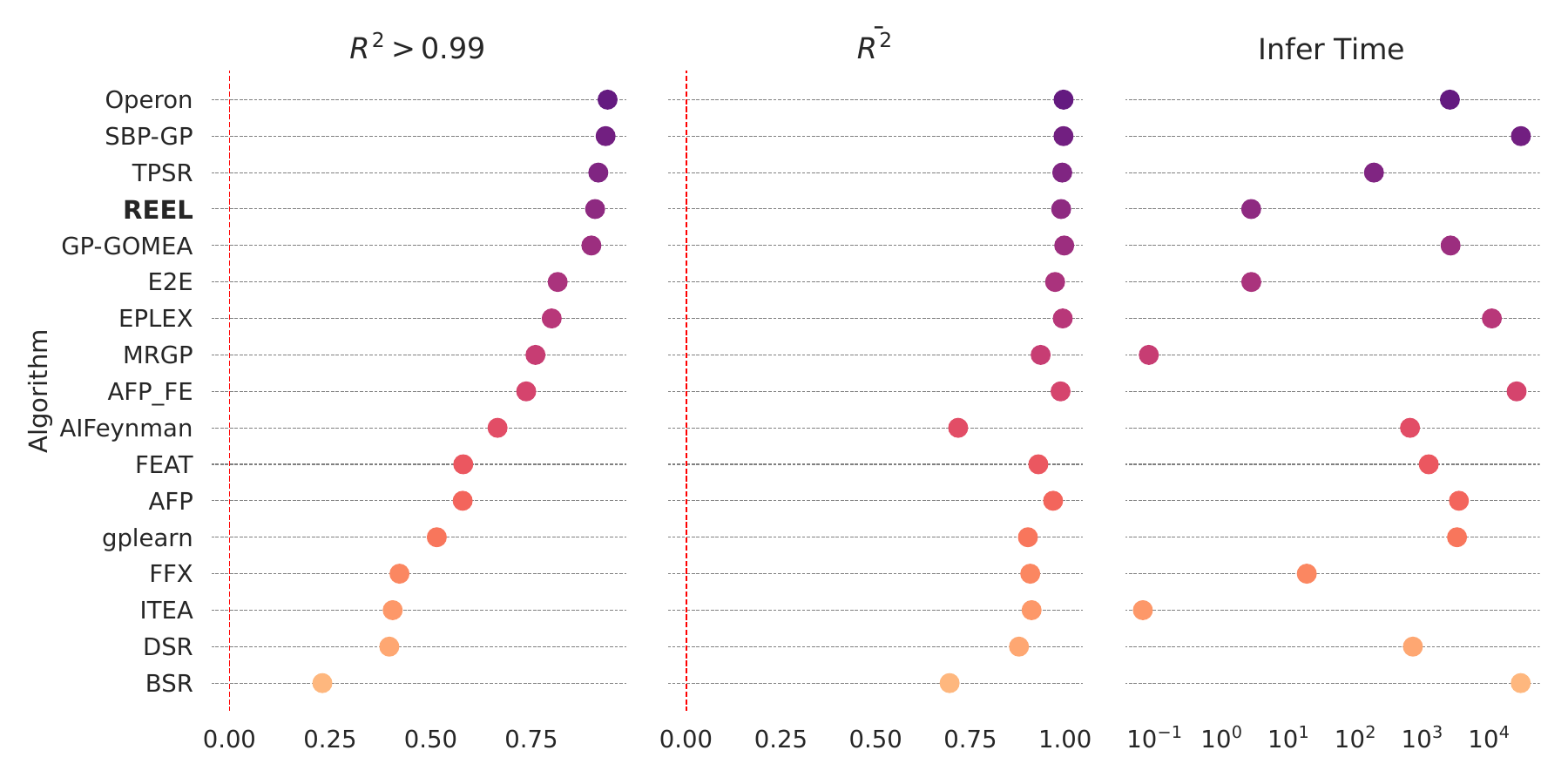}
  }
  \hfill
  \subfigure[Strogatz]{
    \includegraphics[width=0.9\textwidth]{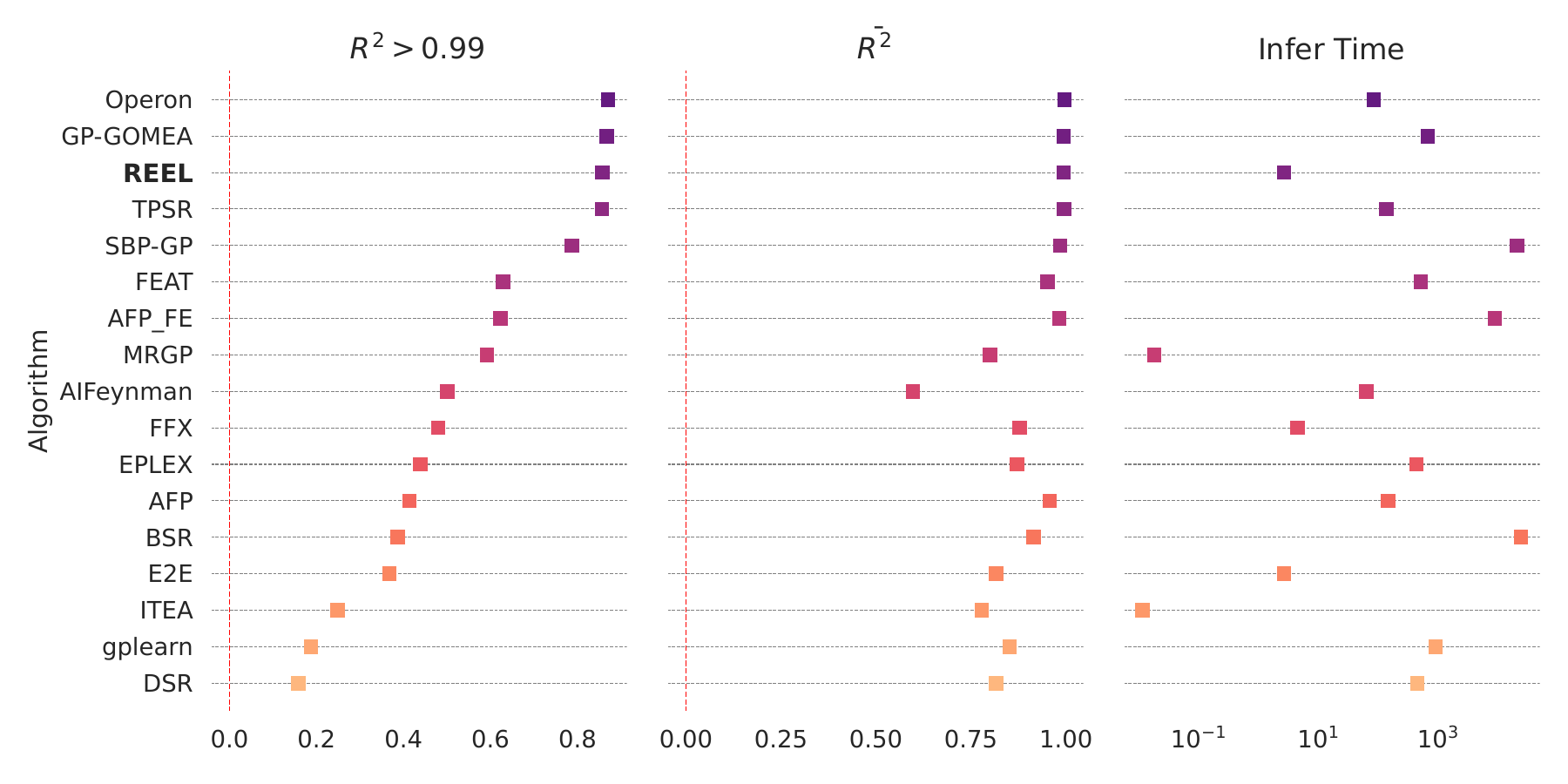}
  }
  \caption{Performance comparison of {\model} and all SRBench algorithms for Feynman and Strogatz datasets. For each dataset, the performance is sorted by the proportion of $R^2 > 0.99$. Our approach presents strong accuracy-speed tradeoffs.}
  \label{fig:3in1}
\end{figure}

\vspace{-0.2cm}
\subsection{Validating the Effectiveness of Reinforcement Learning Feedback}
\vspace{-0.2cm}
To verify the effectiveness of reinforcement learning (RL) in enhancing a foundation model, we conducted experiments comparing the original E2E model with our RL-enhanced variant (REEL) across three benchmark datasets: Feynman, Strogatz, and Black-box. We evaluate four configurations: E2E-Beam and E2E-Sampling, which represent the baseline performance of the pretrained model using beam search and stochastic sampling, respectively; and REEL-Beam and REEL-Sampling, which apply reinforcement learning finetuning under the same decoding strategies.
\textbf{Table~\ref{exp:rl_comparison}} reveals a consistent and substantial performance gain after reinforcement learning. On all datasets, REEL-Sampling achieves the highest scores, particularly excelling in the proportion of equations with $R^2 > 0.99$, improving from 0.815 to 0.908 on Feynman, from 0.357 to 0.857 on Strogatz, and from 0.000 to 0.088 on the challenging Black-box dataset. Similarly, the average $\bar{R^2}$ scores improve significantly. These improvements are also observed in REEL-Beam. We attribute these gains to the advantage-driven finetuning strategy, which effectively aligns equation generation with downstream performance rewards. Reinforcement learning allows the model to explore diverse generation paths and adaptively favor those leading to high-quality symbolic equations. The results support that reinforcement learning boosts the foundation model’s ability to generate accurate equations, validating its effectiveness in equation generation tasks across diverse domains.

\vspace{-0.2cm}
\subsection{Benchmarking {\model} Against Existing Baselines}
\vspace{-0.2cm}
To evaluate the effectiveness of our proposed method {\model} in Data2Eqn tasks, we conduct a comprehensive comparison with a range of baselines from the SRBench on two widely-used datasets: Feynman and Strogatz. Each method is assessed using three metrics: (1) the proportion of test cases with $R^2 > 0.99$, (2) the average $R^2$, and (3) the inference time (log scale). The algorithms are ranked by their $R^2 > 0.99$ proportion. The evaluation is intended to verify whether {\model} can achieve strong equation discovery performance while maintaining practical efficiency.
\textbf{Figure~\ref{fig:3in1}} shows that {\model} achieves near-optimal accuracy in terms of both the strict $R^2 > 0.99$ threshold and average $R^2$. Compared with the top-performing GP-based methods such as Operon and GP-GOMEA, {\model}’s performance differs by only one or two equations in the $R^2 > 0.99$ category, while maintaining nearly identical average $R^2$. This demonstrates that {\model} effectively discovers high-quality symbolic equations with minimal loss in precision.
Notably, {\model} achieves significantly lower inference time compared to many baselines. While traditional approaches typically require costly search procedures during each inference phase, {\model} benefits from its foundation knowledge and reinforcement learning training, allowing it to converge once and make predictions efficiently during deployment. This inference-time efficiency is especially valuable in real-world applications that demand repeated and rapid symbolic predictions. By shifting computational complexity to the training stage, {\model} offers a scalable and practical solution with strong accuracy-speed tradeoffs.
In summary, these experiments demonstrate that {\model} not only achieves competitive performance but also provides substantial gains in inference efficiency, making it well-suited for high-frequency and large-scale scientific discovery tasks.

\vspace{-0.2cm}
\subsection{Robustness Evaluation of {\model} under Varying Noise}
\vspace{-0.2cm}
\begin{wrapfigure}{r}{0.5\textwidth}
    \centering
    \vspace{-0.6cm}
    \subfigure[Feynman]{\includegraphics[width=0.24\textwidth]{{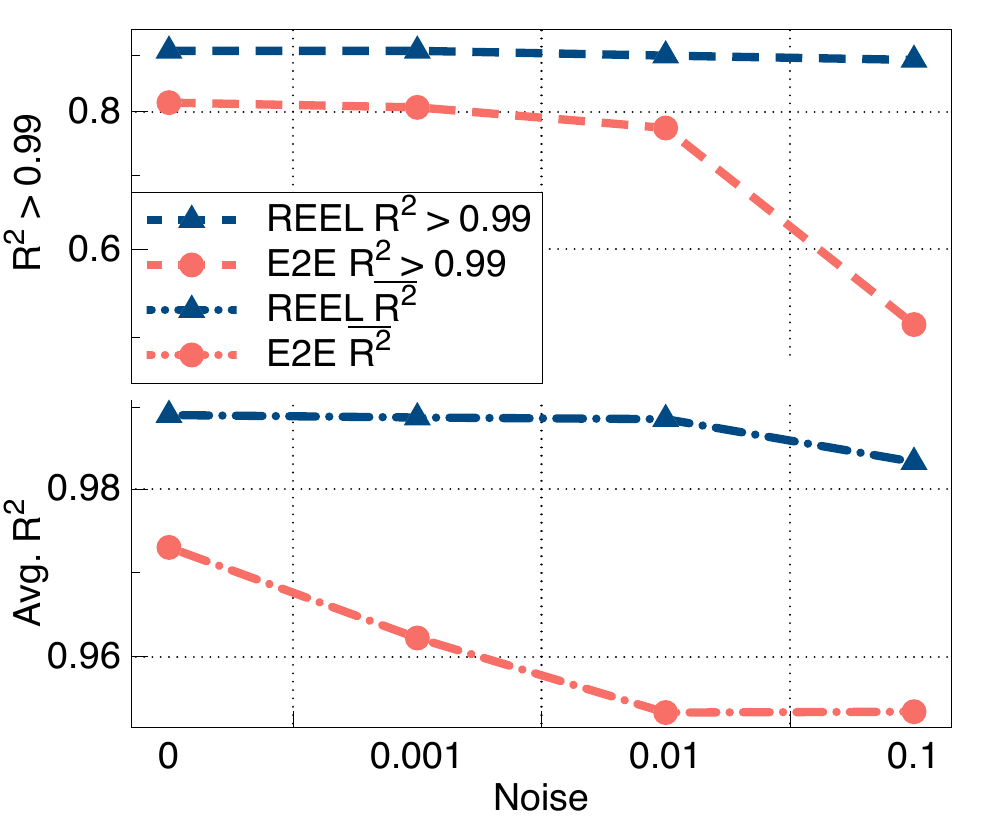}}}
    \subfigure[Strogatz]{\includegraphics[width=0.24\textwidth]{{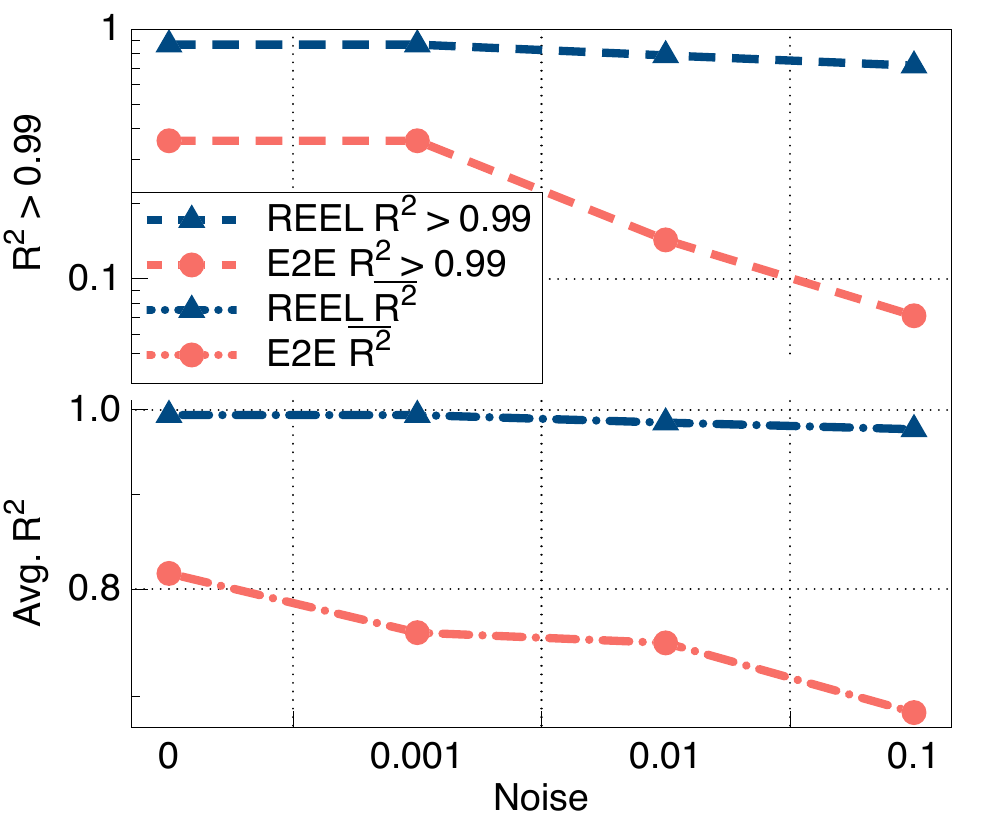}}}
    \caption{Robustness to noise, where the performances are shown for various target noise levels.}
    \label{fig:robust}
\end{wrapfigure}
To evaluate the robustness of {\model} to noise, we conduct experiments by injecting varying levels of Gaussian noise into the values $y$ of the input tables during training. These noisy outputs affect the reward signal, simulating a realistic setting where measurements may be imprecise or corrupted. We compare {\model} against the E2E baseline model under different noise levels on both Feynman and Strogatz datasets.
\textbf{Figure~\ref{fig:robust}} shows that {\model} consistently maintains high performance across all noise levels and remains stable as noise increases, whereas the performance of E2E degrades significantly, especially under moderate to high noise. For example, at the highest noise level 0.1, {\model} still retains a high proportion of accurate equations, while E2E's discovery capability drops sharply.
These findings highlight {\model}’s strong robustness to noise, which we attribute to its reinforcement learning-based optimization framework. Rather than directly fitting noisy outputs, {\model} learns symbolic equations by optimizing expected long-term rewards across multiple equation samples. This policy-based learning paradigm inherently smooths over local noise fluctuations, making it less sensitive to individual corruptions in training data compared to pointwise supervised losses.
In summary, {\model} exhibits excellent resilience under noise, confirming its potential in real-world applications where data quality is imperfect. Its ability to maintain symbolic accuracy under adverse conditions further validates the strength of its reward-driven training strategy.

\noindent To make the experiments more convincing, we deploy complexity and scalability analysis, parameter sensitivity study, and case study in \textbf{Appendix~\ref{appendix:comp},~\ref{appendix:param},~\ref{appendix:case}}, respectively.
\vspace{-0.2cm}
\section{Related Work}
\vspace{-0.2cm}
Symbolic Regression (SR) aims to derive optimal mathematical equations from observed values. Genetic Programming (GP)\citep{GP,GP2,GP3,GP4,GP5} is a classical SR approach that iteratively refines expressions through crossover and mutation. However, GP methods are often computationally expensive and prone to local optima. To improve efficiency, later works have integrated Monte Carlo Tree Search (MCTS)\citep{sun2022symbolic} and reinforcement learning~\citep{mundhenk2021symbolic} to accelerate expression search. Some approaches such as AI Feynman~\citep{udrescu2020ai} incorporate domain knowledge (e.g., physics priors) to constrain the search space, yet they still struggle to generalize across datasets due to a lack of reusable prior knowledge.
The introduction of deep learning has enabled SR models to generate expressions with improved fluency and scalability~\citep{petersen2019deep,kim2020integration,zhang2023deep,d2022deep}, and combining GP with deep networks has further reduced structural search complexity~\citep{cranmer2020discovering}. Recently, Transformer-based and generative models have brought SR closer to pretraining and transfer paradigms. These models learn to generate symbolic equations directly from input-output data pairs, using encoder-decoder architectures originally designed for text generation~\citep{kamienny2022end}. However, their token-level training objectives often neglect mathematical semantics, leading to numerically inaccurate or redundant expressions.
From a broader perspective, symbolic regression can be viewed as a form of data-centric AI—it learns interpretable models directly from observed values without strong priors on functional form. In parallel, recent works in data-centric AI~\citep{wang2025towards,ying2025survey} also emphasize the role of data in model performance, but from a complementary direction: rather than generating symbolic mappings, they optimize the data itself—via feature selection~\citep{ying2024feature,gong2025neuro,wang2024knockoff,ying2024revolutionizing} or transformation~\citep{ying2024unsupervised,ying2023self,ying2024topology,gong2025evolutionary,hu2024reinforcement,gong2025unsupervised}—to boost learning effectiveness. While the modeling goals differ, both SR and data-centric AI adopt a shared architectural backbone—transformer-based encoder-decoders—and are driven by the same principle: extracting structure from data. In this light, our reinforcement adaptation approach aligns these two views by optimizing symbolic generation (as in SR) with feedback derived from data quality (as in data-centric AI), bridging the gap between structure discovery and data optimization.
\vspace{-0.3cm}
\section{Limitations}
\vspace{-0.2cm}
Despite the effectiveness of {\model}, our approach inherits certain limitations from the underlying foundation model. The pretrained model imposes strict constraints on input dimensionality, typically requiring the input table to contain fewer than ten variables. This is because transforming high-dimensional tabular data into token sequences results in excessively long input sequences, which are difficult to handle efficiently. Moreover, as dimensionality increases, the corresponding symbolic expressions become more complex and harder to optimize, leading to longer and less tractable generation sequences on the decoder side. Therefore, we follow the same dimensionality constraint as used in prior foundation model settings.
\vspace{-0.3cm}
\section{Conclusion}
\vspace{-0.2cm}
In this work, we propose {\model}, a reinforcement learning-based finetuning framework that improves the domain adaptability and numerical accuracy of foundation models for the data-to-equation (Data2Eqn) task. By optimizing generation policies using reward signals derived from downstream equation fitness, {\model} addresses the limitations of token-level objectives and enhances the model’s ability to recover meaningful mathematical expressions. Extensive experiments demonstrate that {\model} improves accuracy and robustness to noise across multiple benchmarks. These results suggest that reinforcement-guided finetuning is a practical and effective strategy for adapting pretrained models to domain-specific data-to-equation tasks.
\small
\bibliographystyle{plainnat}
\bibliography{ref.bib}
\appendix

\section{Algorithm: Reinforcement Finetuning for Learning-centric Foundation Model Adaptation}
\label{appendix:algo}
\begin{algorithm}[th]
\caption{Two-Loop Reinforcement Finetuning Framework}
\KwIn{Sampled subsets $\mathcal{T} = \{(\mathbf{X}_i, y_i)\}_{i=1}^N$, Pretrained model $\pi_{\text{old}}$, reward function $R(\cdot)$}
\KwOut{Finetuned policy $\pi_\theta$}

Initialize training policy: $\pi_\theta \leftarrow \pi^{old}_\theta$\;

\For{\textbf{Outer Loop}}{

    Select a sampled subset $(\mathbf{X}_i, y_i)$ as input;

    \For{\textbf{Inner Loop}}{
        Encode input to obtain contextual embedding $\mathbf{e}$\;
        
        Generate token sequence $a_{1:T} \sim \pi_\theta$\;

        Compute reward $r \leftarrow R(a_{1:T}, (\mathbf{X}_i, y_i))$\tcp*[r]{Environment feedback}
        
        Compute importance ratio:\quad
        $\rho = \exp (\frac{1}{T} \Sigma_{t=1}^T \log \frac{\pi_\theta(a_t|s_t)}{\pi_{\theta}^{old}(a_t|s_t)})$\;
    
        Compute clipped surrogate loss:\quad
        $\mathcal{L}_{\text{clip}} = -\min(\rho \cdot r,\ \text{clip}(\rho, 1-\epsilon, 1+\epsilon) \cdot r)$\;

        Compute step-wise KL divergence:\quad
        $\mathcal{L}_{kl} = \frac{1}{T} \sum_{t=1}^T \sum_{a \in \mathcal{V}} \pi^{old}_\theta(a \mid s_t) \left[\log \pi^{old}_\theta(a \mid s_t) - \log \pi_\theta(a \mid s_t)\right]$\;

        Compute total loss:\quad $\mathcal{L} = \mathcal{L}_{\text{clip}} + \beta \cdot \mathcal{L}_{kl}$\;
    
        Update $\pi_\theta$ via gradient descent using $\nabla_\theta \mathcal{L}$
    }
}
\Return{$\pi_\theta$}
\label{alg:twoloop}
\end{algorithm}

\section{Experimental Settings}
\label{appendix:exp_set}
\noindent\textbf{Evaluation Metrics.} 
1) \textbf{$\bm{R^2}$ score:} $R^2 = 1-\frac{\Sigma_{i=1}^n(\Tilde{y}_i-y_i)^2}{\Sigma_{i=1}^n(\Tilde{y}_i-\bar{y})^2}$, where $R^2$ measure the fitness accuracy of the generated equations. 
2) \textbf{$\bm{R^2 > 0.99}$:} For a given domain-specific dataset (e.g., Feynman), $R^2 > 0.99$ denotes the proportion of equations achieving an $R^2$ score larger than 0.99. For example, if 80 out of 119 equations in the Feynman datasets reach this threshold, then $R^2 > 0.99 = \frac{80}{119} = 0.672$. 
3) \textbf{Data splitting:} We split the datapoints of each equation into training and testing datapoints at a ratio of 75\% and 25\%, respectively. 
The finetuning datasets are sampled from the training datapoints, while the test datapoints are used to evaluate model performance.

\noindent\textbf{Datasets.} We used three widely used standard benchmark datasets:
\textbf{1) Feynman:} The Feynman dataset comprises 119 physical equations of the real world~\cite{udrescu2020aifeynman20paretooptimal}. For each equation, the input pairs $(\mathbf{X},y)$ are sampled by Penn Machine Learning Benchmark (PMLB)~\cite{olson2017pmlblargebenchmarksuite} and are utilized in SRBench~\cite{lacava2021contemporarysymbolicregressionmethods}. Each equation has an input dimensionality (a.k.a., number of features) of less than 10.
\textbf{2) Strogatz:} The Strogatz dataset is constructed based on nonlinear dynamics and consists of 14 equations~\cite{la2016inference}. It is widely used in symbolic regression research. Each equation in the dataset corresponds to an input dimensionality of 2.
\textbf{3) Black-box:} The black-box dataset is sourced from PMLB and has been studied in multiple symbolic regression baselines. Since the pretraining of the foundation model strictly limits the input feature dimensionality to 10 or less, we select only those black-box datasets that meet this criterion. Finally, the dataset consists of a total of 57 equations.

\noindent\textbf{Implementation Details.}
Our model implementation leverages the state-of-the-art open-source End-to-End (E2E) model~\cite{kamienny2022end} as the pretrained transformer backbone. This selection is due to the public availability of E2E's model architecture, pretrained weights, and output logits through the Facebook symbolic regression library and its associated repository\footnote{\url{https://github.com/facebookresearch/symbolicregression}}. Our reinforcement learning-based finetuning code based on E2E is publicly released.
\textbf{1) Backbone: Numerical Encoder and Decoder:} We reload the E2E pretrained weights and ensure the hyperparameters remain consistent with the original model. The model's maximum sequence length is set to 200.
\textbf{2) Finetuning data sampling:} To construct the finetuning dataset, we sample 128 bags of 200 input points (a.k.a., rows) for each equation from the training data. This setup aligns with the pretraining data of the Transformer backbone E2E, which is pretrained on large datasets with input points $\leq$ 200.
\textbf{3) Reinforcement learning-based finetuning:} The learning rate is set to 5e-5 using the AdamW optimizer. We set the batch size to 64. During the finetuning stage, the RL update is performed for 10 training epochs. For policy regularization, we include a KL-divergence penalty with coefficient $\beta = 0.2$ and clip the policy update ratio using $\epsilon = 0.2$. In addition, equations can vary significantly in difficulty—some are easy to recover, while others are structurally complex. Mixing such equations in the same training batch can lead to reward signal divergence and training instability. To mitigate this, we train on each equation instance separately to ensure reward consistency and optimization stability.
\textbf{4) Environmental settings:} All experiments are conducted on the Ubuntu 22.04.3 LTS operating system, Intel(R) Xeon(R) w9-3475X CPU@ 4800MHz, and 1 way RTX A6000 and 48GB of RAM, with the framework of Python 3.11.4 and PyTorch 2.5.1.

\noindent\textbf{Baselines.}
We compare {\model} with the backbone E2E and various SRBench algorithms\footnote{\url{https://github.com/cavalab/srbench}} to evaluate our proposed method.
\textbf{1) End-to-end symbolic regression with transformers (E2E):} E2E~\cite{kamienny2022end} trains Transformers to directly generate full symbolic expressions, including constants, from input data in a single pass, offering high accuracy and orders-of-magnitude faster inference than GP-based methods.
\textbf{2) Operon:} Operon~\cite{GP} is a fast C++ genetic programming framework for symbolic regression with linear tree encoding and parallel evolution.
\textbf{3) Semantic Backpropagation GP (SBP-GP):} SBP-GP~\cite{SBP-GP} uses semantic backpropagation to guide subtree replacement in genetic programming, and is enhanced with linear scaling to improve symbolic regression accuracy and generalization.
\textbf{4) Transformer-based Planning for Symbolic Regression (TPSR):} TPSR~\cite{shojaee2023transformer} integrates Monte Carlo Tree Search with pretrained Transformer SR models to guide equation generation using feedback on accuracy and complexity, improving both fitting and interpretability.
\textbf{5) GP-based Gene-pool Optimal Mixing Evolutionary Algorithm (GP-GOMEA):} GP-GOMEA~\cite{GP-GOMEA} applies gene-pool optimal mixing to genetic programming, using linkage learning and entropy-based building-block identification to improve scalability and solution compactness.
\textbf{6) Epsilon-Lexicase Selection (EPLEX):} EPLEX~\cite{EPLEX} is a parent selection method for symbolic regression that relaxes elitism in lexicase selection using adaptive $\epsilon$ thresholds, improving performance in continuous domains.
\textbf{7) Multiple Regression GP (MRGP):} MRGP~\cite{MRGP} enhances symbolic regression by linearly combining program subexpressions via multiple regression, using them as features to improve predictive accuracy.
\textbf{8) Age-Fitness Pareto Optimization (AFP):} AFP~\cite{AFP} maintains evolutionary diversity by selecting individuals on a Pareto front of fitness and age, helping avoid premature convergence in symbolic regression.
\textbf{9) Age-Fitness Pareto Optimization with Co-evolved Fitness Predictors 
 (AFP\_FE):} AFP\_FP~\cite{AFP} extends AFP optimization by co-evolving fitness predictors to accelerate evaluation, enabling efficient symbolic regression on large datasets.
\textbf{10) AIFeynman:} AIFeynman~\cite{udrescu2020aifeynman20paretooptimal} combines neural networks, graph modularity discovery, and Pareto-optimal pruning to robustly recover interpretable symbolic equations from noisy data.
\textbf{11) Feature Engineering Automation Tool (FEAT):} FEAT~\cite{FEAT} evolves networks of syntax trees with differentiable weights to learn compact and interpretable representations for symbolic regression.
\textbf{12) gplearn:} gplearn performs symbolic regression using genetic programming, providing interpretable models through a scikit-learn-style interface.
\textbf{13) Fast Function Extraction (FFX):} FFX~\cite{FFX} is a deterministic symbolic regression method that combines massive basis function enumeration with pathwise regularized learning to efficiently extract compact, interpretable models.
\textbf{14) Interaction-Transformation Evolutionary Algorithm (ITEA):} ITEA~\cite{ITEA} evolves structured affine combinations of nonlinear feature interactions using mutation-only genetic search, enabling interpretable and scalable symbolic regression.
\textbf{15) Deep Symbolic Regression (DSR):} DSR~\cite{petersen2019deep} trains an autoregressive RNN with a risk-seeking policy gradient to generate symbolic expressions that maximize best-case performance in symbolic regression.
\textbf{16) Bayesian Symbolic Regression (BSR):} BSR~\cite{BSR} frames symbolic regression in a Bayesian framework by linearly combining symbolic trees sampled via MCMC, enabling interpretable expressions and incorporation of prior knowledge.

\section{Complexity and Scalability Analysis of {\model} in Real-world Dataset}
\label{appendix:comp}
\begin{figure}[t]
  \centering
  \subfigure[Time Complexity Check]{
    \includegraphics[width=0.3\textwidth]{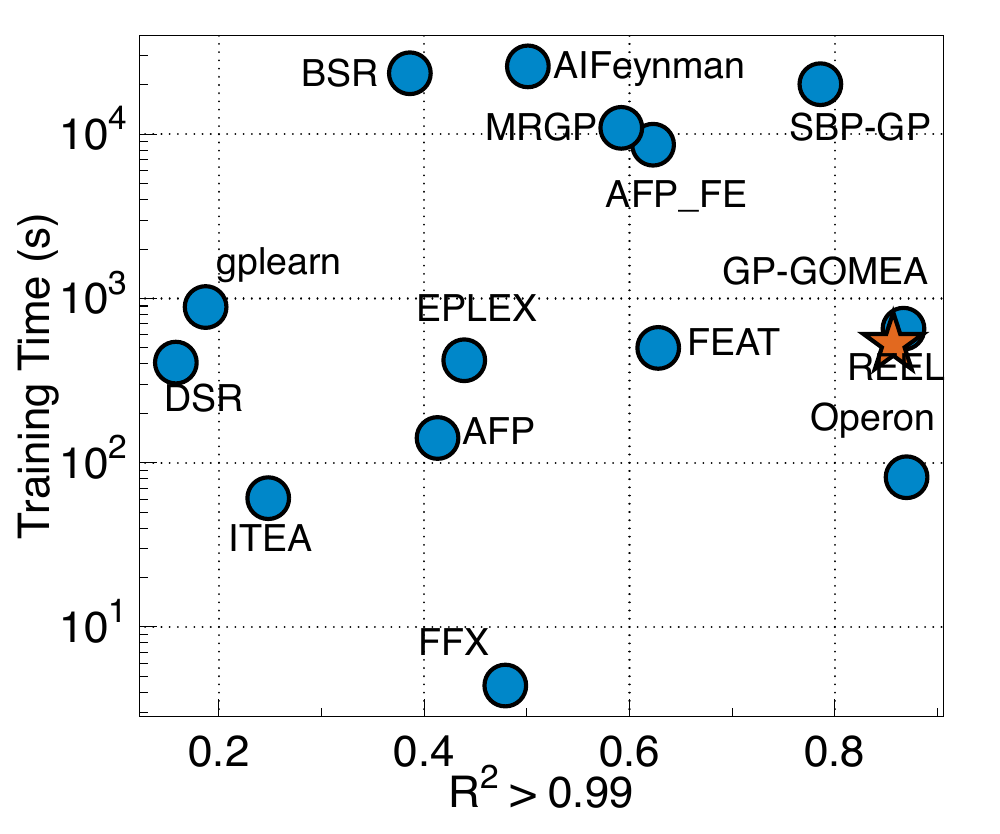}
  }
  \hfill
  \subfigure[Model Complexity Check]{
    \includegraphics[width=0.3\textwidth]{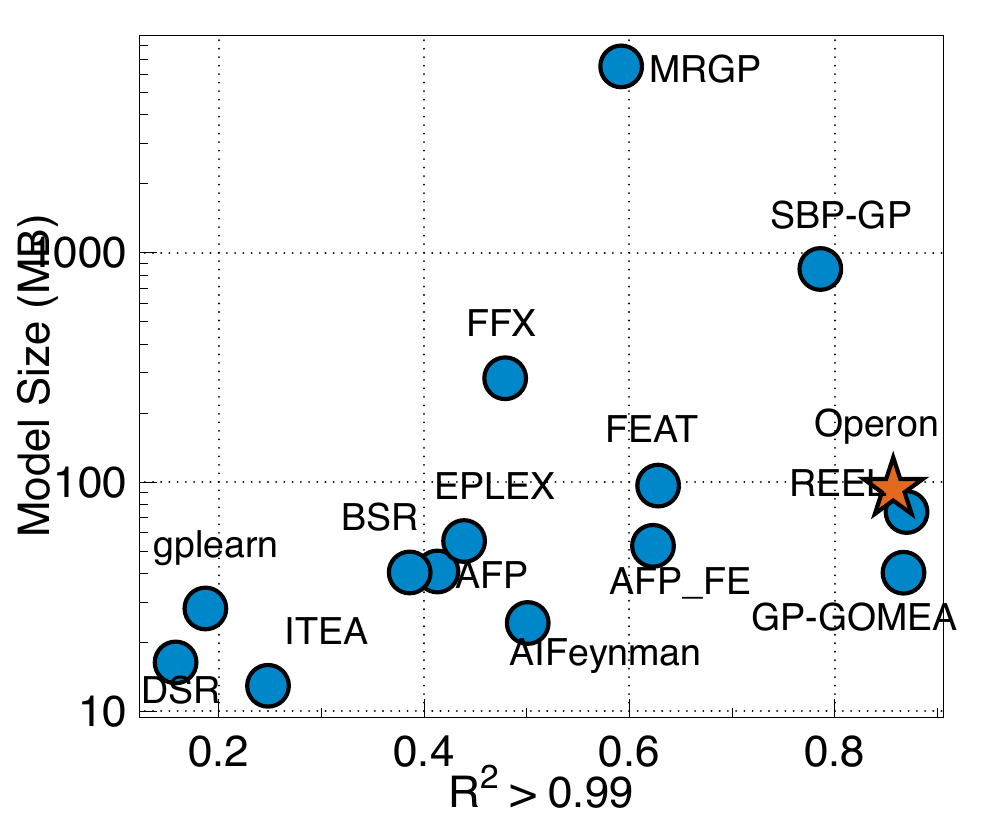}
  }
  \hfill
  \subfigure[Scalability Check]{
    \includegraphics[width=0.3\textwidth]{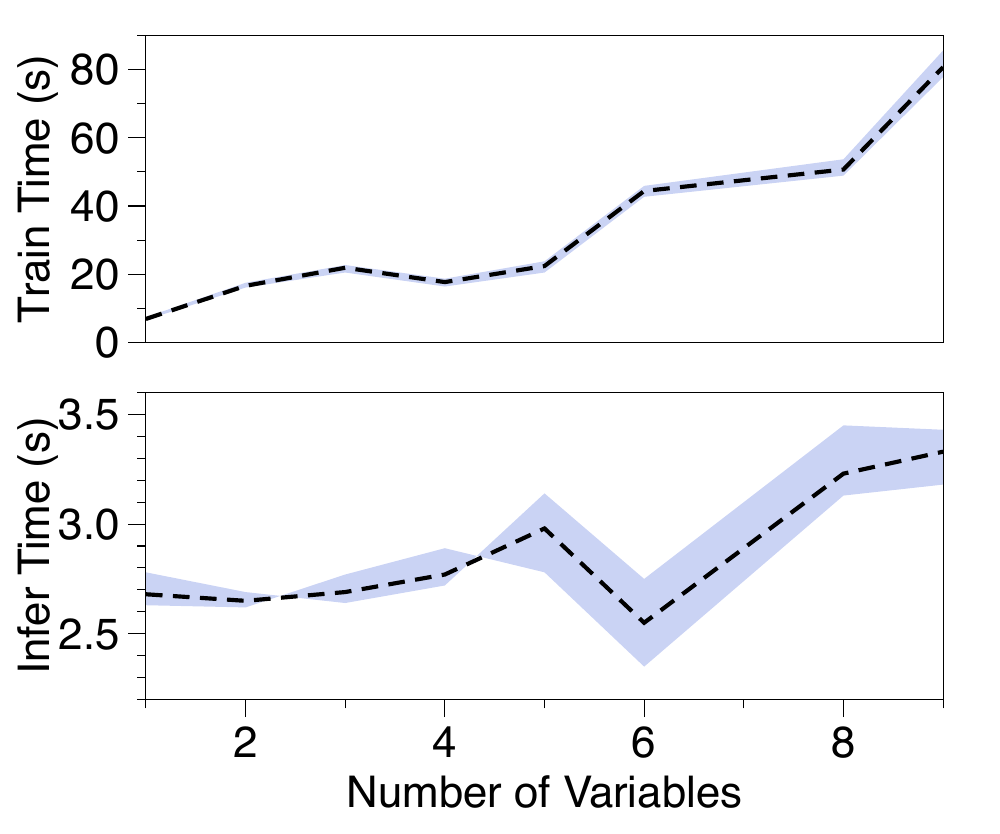}
  }
  \caption{Complexity and Scalability Analysis of {\model}. (a) Comparison of training time on the Strogatz dataset between {\model} and the baselines; (b) Comparison of model size between {\model} and the baselines; (c) Analysis of how training time per epoch and prediction time change as the number of variables (features) in the dataset increases.}
  \label{complexity}
\end{figure}

This experiment aims to evaluate the efficiency and scalability of our proposed method {\model} in terms of computational cost and model complexity.
1) \textbf{Figure~\ref{complexity}(a)} shows that models such as BSR, MRGP, and AIFeynman require substantial computational resources but fail to achieve high accuracy. FFX exhibits extremely fast training but at the cost of low accuracy. In contrast, {\model} achieves high accuracy while maintaining relatively short training time. Notably, once the model converged, {\model} demonstrates a much lower inference time than other baselines, as shown in \textbf{Figure~\ref{fig:3in1}(b)}, highlighting the efficiency of our finetuning mechanism in reducing computational overhead without compromising performance.
2) \textbf{Figure~\ref{complexity}(b)} explores the relationship between model size and accuracy. Models like SBP-GP attain high accuracy at the cost of large storage requirements, while others like ITEA, DSR, and gplearn sacrifice accuracy to keep the model compact. In contrast, {\model} achieves an effective balance: it maintains a smaller model size than most high-accuracy baselines while still delivering strong performance.
3) \textbf{Figure~\ref{complexity}(c)} evaluates scalability on a real-world dataset from the physics domain (Feynman benchmark). As the number of variables increases, training time per epoch grows—remaining relatively stable in low-dimensional settings but exhibiting nonlinear growth in higher dimensions. This behavior likely results from the longer token sequences generated when converting high-dimensional tables into tokens and complexity of equations. Nevertheless, the runtime remains within an acceptable range. Prediction time, on the other hand, remains consistently low and stable across different input sizes. This efficiency benefits from the design of {\model}, which maps tabular input into fixed-size vector representations and then reconstructs them into equations, ensuring that prediction speed is unaffected by input dimensionality.
These results demonstrate that {\model} not only offers a favorable accuracy-efficiency trade-off but also scales well with increasing input dimensionality. Its finetuning mechanism and compact model design enable efficient learning and fast inference, making it practical for real-world symbolic regression tasks.

\section{Hyperparameter Sensitivity Study}
\label{appendix:param}
\begin{wrapfigure}{r}{0.5\textwidth}
    \vspace{-0.2cm}
    \centering
    \subfigure[Sensitivity of $\beta$]{\includegraphics[width=0.24\textwidth]{{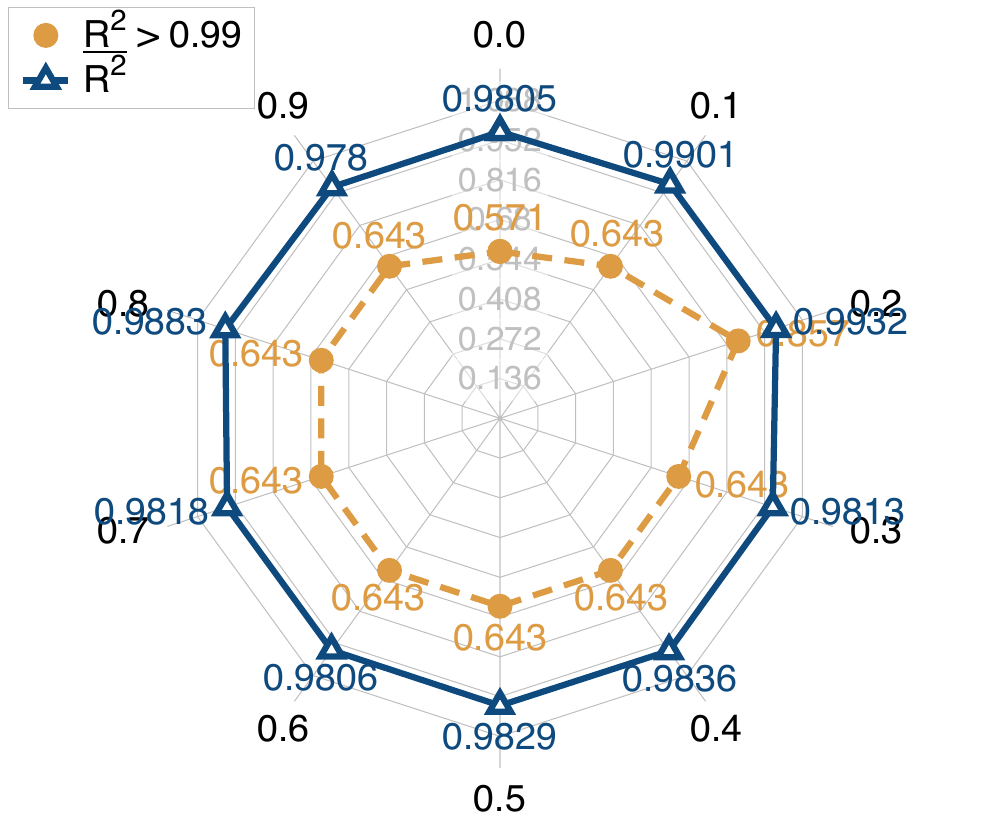}}}
    \subfigure[Sensitivity of $\epsilon$]{\includegraphics[width=0.24\textwidth]{{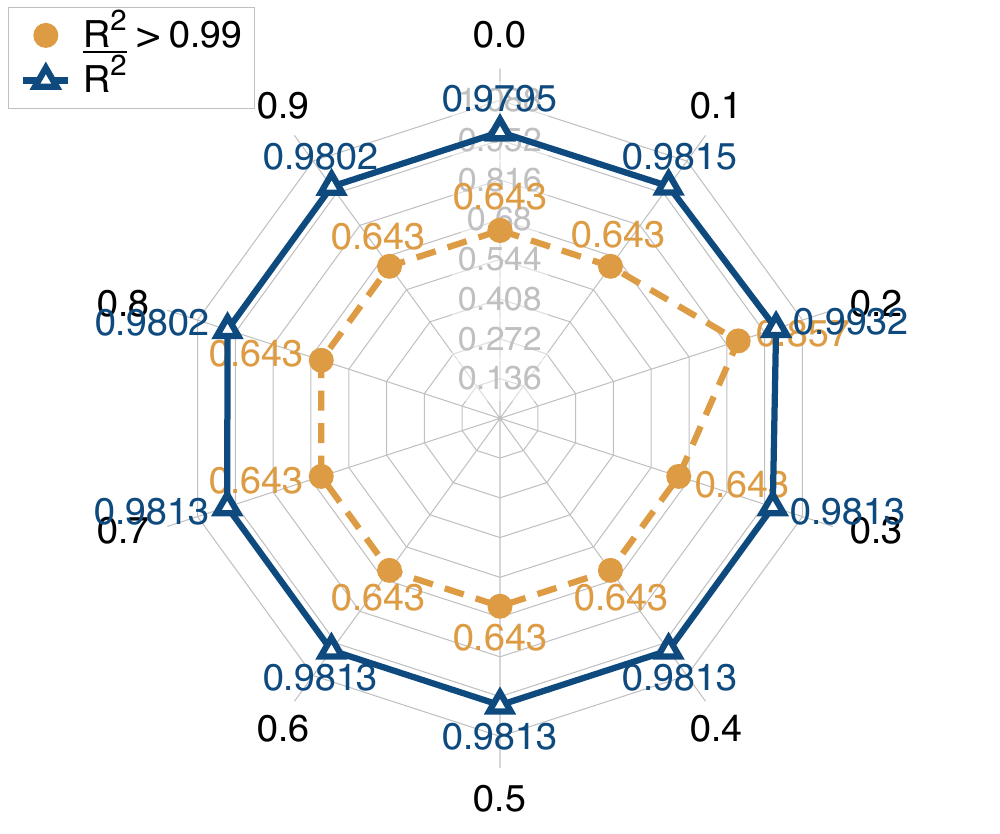}}}
    \caption{Hyperparameter sensitivity analysis. (a) $\beta$ controls how much the updated policy deviates from the pretrained model (via KL divergence), and (b) $\epsilon$ limits the update size of the policy.}
    \label{fig:sense}
\end{wrapfigure}
This experiment investigates the sensitivity of reinforcement learning finetuning to two key hyperparameters: $\beta$, which constrains the KL divergence between the updated policy and the pretrained model, and $\epsilon$, which limits the magnitude of policy updates. \textbf{Figure~\ref{fig:sense} (a) and (b)} illustrate how variations in these parameters affect the performance of the {\model} on the Data-to-Equation task.
1) In \textbf{Figure~\ref{fig:sense} (a)}, we observe that {\model} achieves the best performance when $\beta=0.2$. This indicates that a moderate constraint on policy divergence helps stabilize training and allows for effective exploration. When $\beta$ is set to 0.0, the absence of regularization leads to overly aggressive exploration, causing the model to deviate from the pretrained distribution and resulting in performance degradation. On the other hand, when $\beta$ exceeds 0.2, the performance drops and stabilizes, suggesting that excessive KL penalties overly restrict optimization and prevent the model from escaping suboptimal regions.
2) \textbf{Figure~\ref{fig:sense} (b)} shows the effect of varying $\epsilon$, which controls the allowed extent of policy changes. The best results are obtained at $\epsilon=0.2$, where policy updates are sufficiently flexible for learning while avoiding instability. When $\epsilon$ is too small, updates become overly conservative, limiting the model’s capacity to improve. Conversely, a large $\epsilon$ permits drastic policy shifts, which can introduce high-variance updates and hinder convergence.
These findings highlight the importance of carefully tuning the KL divergence penalty and updating magnitude constraints in reinforcement learning-based finetuning. Properly balancing these two factors ensures stability during training while maintaining sufficient learning capacity. In our setting, $\beta=0.2$ and $\epsilon=0.2$ provide the optimal trade-off, significantly enhancing the quality of the generated equations.

\section{Case Study: Interpretable Equation Discovery in Real-world Datasets}
\label{appendix:case}
\setlength{\tabcolsep}{1.9mm}{
\begin{table*}[th]
\centering
\caption{Case study on the Feynman dataset showcasing equation generation results by E2E and {\model}. {\model} recovers the key structural components of the true physical equations, resulting in significantly improved $R^2$ scores compared to the E2E baseline.}
\vspace{0.2cm}
\begin{tabular}{@{}ccc@{}}
\toprule \toprule
                 & Equation & $R^2$ \\ \midrule
True Equation    & $2x_0(1-\bm{cos(x_1x_2)})$        & -                    \\
E2E Generation   & $Cx_0(Csin(Cx_1+Cx_2+C)+(Cx_1+C)(Cx_2+C))$        & 0.3958               \\
{\model} Generation & $Cx_0(C+C\bm{cos}(Cx_1+Cx_2+\bm{Cx_1x_2}))+C$       & 0.9969               \\ \midrule
True Equation    & $\sqrt{x_0^2+x_1^2-2x_0x_1\bm{cos(x_2-x_3)}} $        & -                    \\
E2E Generation   & $C+C\sqrt{|(x_0+x_1)sin(Cx_0^2)(x_2+x_3)|}$        & 0.9241               \\
{\model} Generation & $C\sqrt{C(x_0+x_1)\bm{cos(Cx_2-Cx_3)}+Cx_0x_1}$        & 0.9727               \\ \midrule
True Equation    & $x_0(\bm{cos(x_1x_2)}+x_3\bm{cos^2(x_1x_2)})$        & -                    \\
E2E Generation   & $Cx_0(|(x_1+x_2)sin(x_1+x_2)|-cx_3)$        & 0.8577               \\
{\model} Generation & $Cx_0(C\bm{cos^2(Cx_1x_2)}+Cx_2\bm{cos(Cx_1x_2)}+ Cx_2^2+Cx_3+C)$        & 0.9517               \\ \midrule
True Equation    & $\bm{x_4x_0x_1(\frac{x_2-x_3}{x_2x_3})}$       & -                    \\
E2E Generation   & $Cx_1(sin(Cx_4)+Cx_4(x_0+x_2)(x_2+x_3)(x_3+x_2^2)$        & 0.8736               \\
{\model} Generation & $\bm{(Cx_0+C)(Cx_1+C)(Cx_4+C)(\frac{C(x_2-x_3)}{x_2x_3})}+C$        & 1.0000               \\ \midrule
True Equation    & $\frac{x_0\bm{sin^2(\frac{x_2x_1}{2})}}{sin^2(\frac{x_1}{2})}$        & -                    \\
E2E Generation   & $Cx_0(cos(cx_2)+C(x_1cos^2(Cx_2))$        & 0.5225               \\
{\model} Generation & $Cx_0(1-\bm{sin^2(Cx_2x_1} + Cx_1^2))$        & 0.9673               \\ \bottomrule \bottomrule
\end{tabular}
\label{exp:case_study}
\end{table*}}
This case study aims to assess the performance of {\model} on a set of challenging equations from the Feynman dataset. These cases were selected to represent difficult scenarios where the target equations involve complex structures such as nested nonlinear functions, multiplicative interactions, and higher-order compositions—settings that often challenge Data2Eqn models. To simplify the notations, we replace the constant with 'C' in the equations.
As shown in \textbf{Table~\ref{exp:case_study}}, the baseline E2E model fails to recover the core mathematical structures of these equations, resulting in low $R^2$ scores. In contrast, {\model} reconstructs key terms such as $x_1x_2$, $\cos^2(x)$, or $\frac{x_2-x_3}{x_2x_3}$, which are essential to the semantic integrity of the target expressions. For instance, in Case 4, REEL precisely reproduces the multiplicative and fractional terms, achieving an exact match with the ground truth ($R^2=1.000$), while E2E generates an overcomplicated but structurally incorrect expression.
The strong performance of {\model} in these hard cases can be attributed to its reward-driven finetuning mechanism, which guides the model to favor equations that fit the data well with mathematical semantics. By iteratively optimizing equation quality based on reward feedback, {\model} avoids overfitting to noisy or spurious patterns.
In summary, this case study demonstrates {\model}’s ability to recover accurate and interpretable symbolic equations under structurally complex and high-difficulty settings, highlighting its potential in scientific reasoning tasks that demand both precision and mathematical semantics.

\section{Broader Impact}
Our work enhances the interpretability and adaptability of foundation models through reinforcement-guided symbolic regression. This approach has the potential to benefit high-stakes domains such as renewable energy forecasting~\citep{huo2025ct}, financial analysis~\citep{huo2025enhancing,li2023sehf}, stock prediction~\citep{202501.1003}, recommender system~\citep{rs1,rs2,rs3,wang2024llm}, and biomedical analysis~\citep{liu2019edta,wang2022successful,li2024sade,wang2024lcmdc,liu2024calorie,liu2024pth}, where transparent and data-aligned models are essential. By aligning generation with data semantics, REEL supports responsible and trustworthy AI deployment, though caution is needed when applying symbolic models in highly noisy or high-dimensional settings.

\end{document}